\documentclass[11pt]{article}

\usepackage{acl}

\usepackage{times}
\usepackage{latexsym}

\usepackage[T1]{fontenc}

\usepackage[utf8]{inputenc}

\usepackage{microtype}

\usepackage{inconsolata}

\usepackage{booktabs}
\usepackage{multirow}
\usepackage{multicol}
\usepackage{makecell}
\usepackage{enumitem}
\usepackage{balance}

\usepackage{amsmath} 
\usepackage{amssymb} 
\usepackage{graphicx}

\title{HEAL: Hindsight Entropy-Assisted Learning for Reasoning Distillation}

\author{
  \textbf{Wenjing Zhang}$^{1,2}$\thanks{\ \ Equal contribution.} \quad
  \textbf{Jiangze Yan}$^{1,2}$\footnotemark[1]  \quad
  \textbf{Jieyun Huang}$^{1,2}$\footnotemark[1]  \quad
  \textbf{Yi Shen}$^{1,2}$\thanks{\ \ Corresponding authors.} \quad \textbf{Shuming Shi}$^{1,2}$ \quad \\
  \textbf{Ping Chen}$^{1,2}$  \quad 
  \textbf{Ning Wang}$^{1,2}$ \quad 
  \textbf{Zhaoxiang Liu}$^{1,2}$ \footnotemark[2] \quad
  \textbf{Kai Wang}$^{1,2}$  \quad 
  \textbf{Shiguo Lian}$^{1,2}$\footnotemark[2]\\[2mm] 
$^{1}$ Data Science \& AI Research Institute, China Unicom \\
$^{2}$ Unicom Data Intelligence, China Unicom \\
\texttt{\{zhangwj1503, sheny73, lzx178, liansg\}@chinaunicom.cn}
}

\begin{document}
\maketitle
\begin{abstract}


Distilling reasoning capabilities from Large Reasoning Models (LRMs) into smaller models is typically constrained by the limitations of rejection sampling. Standard methods treat the teacher as a static filter, discarding complex "corner-case" problems where the teacher fails to explore valid solutions independently, thereby creating an artificial "Teacher Ceiling" for the student. In this work, we propose Hindsight Entropy-Assisted Learning (HEAL), an RL-free framework designed to bridge this reasoning gap. Drawing on the educational theory of the Zone of Proximal Development (ZPD), HEAL synergizes three core modules: (1) Guided Entropy-Assisted Repair (GEAR), an active intervention mechanism that detects critical reasoning breakpoints via entropy dynamics and injects targeted hindsight hints to repair broken trajectories; (2) Perplexity-Uncertainty Ratio Estimator (PURE), a ratio-based filtering heuristic that reduces high-anomaly shortcut-like rationales; and (3) Progressive Answer-guided Curriculum Evolution (PACE), a three-stage distillation strategy that organizes training from foundational alignment to hard-case adaptation. Extensive experiments on multiple benchmarks demonstrate that HEAL significantly outperforms traditional SFT distillation and other baselines. 


\end{abstract}

\section{Introduction}

Large reasoning models (LRMs) such as OpenAI-o1 \cite{OpenAI2024o1} and DeepSeek-R1 \cite{guo2025deepseek} have demonstrated strong performance on complex reasoning tasks, including mathematical problem-solving \cite{AIME24}, programming challenges \cite{jain2024livecodebench}, and scientific QA \cite{rein2024gpqa}. Current approaches to constructing LRMs mainly fall into Reinforcement Learning (RL) and Knowledge Distillation \cite{chen2025towards}. While RL can incentivize models to refine their search within valid solution spaces, it often suffers from instability and incoherent outputs in smaller-scale settings \cite{yue2025does}. Distillation offers a more stable and controllable alternative for small models \cite{guo2025deepseek}, enabling them to inherit advanced reasoning patterns that may lie beyond their autonomous exploration capabilities \cite{hu2025distillation, kim2025reinforcement}.

\begin{figure}[t]
\centering
\includegraphics[width=0.46\textwidth]{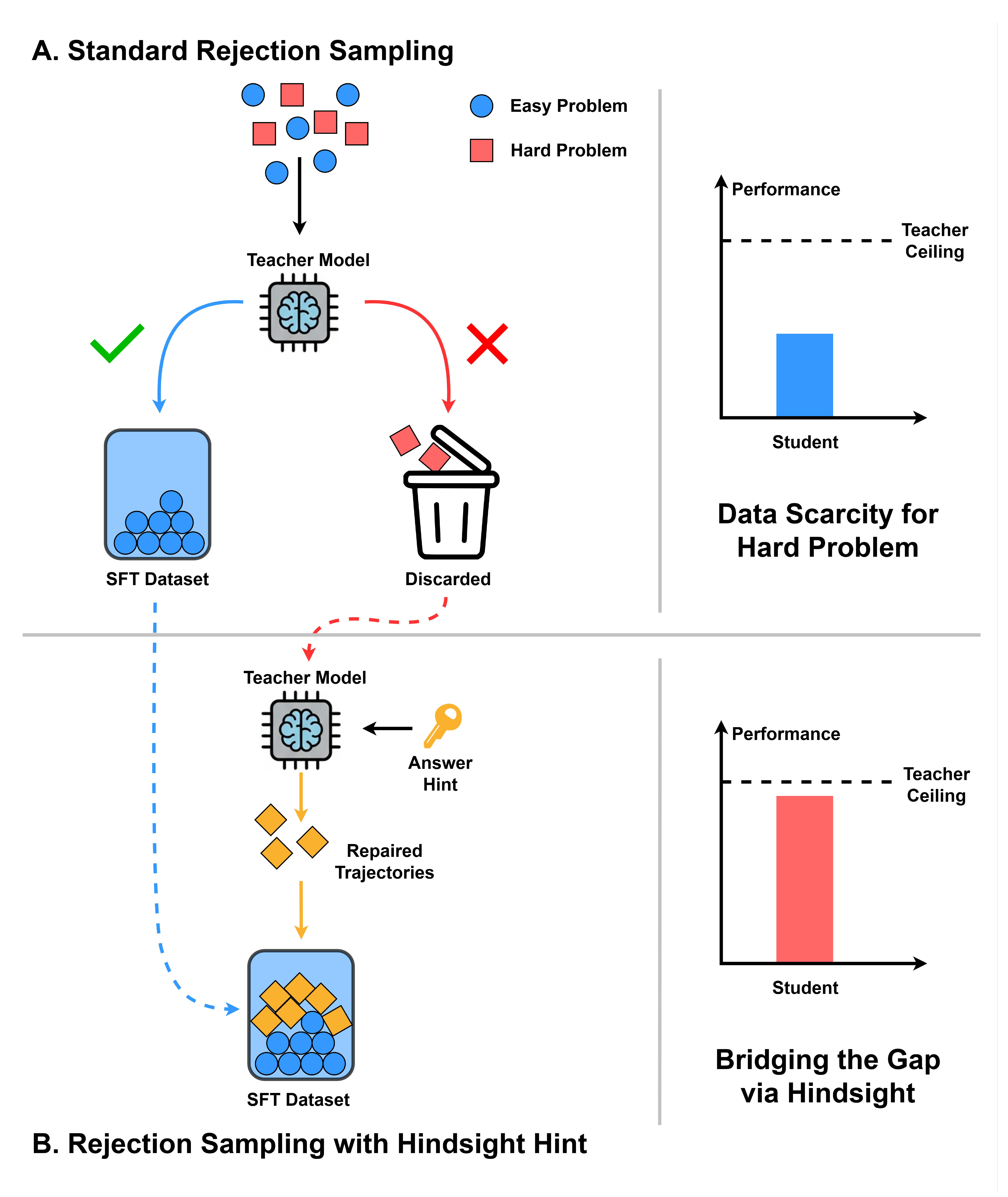}
\caption{(A) The “Teacher Ceiling” challenge of standard rejection sampling for reasoning distillation. (B) Hindsight prompts with hints help recover answer-consistent trajectories for difficult problems, thereby approaching the “Teacher Ceiling”.} 
\label{fig:intro}
\end{figure}

The standard paradigm for reasoning distillation relies on rejection sampling: a teacher model generates multiple reasoning trajectories for each problem, and only trajectories with correct final answers are retained for student training. As shown in Figure~\ref{fig:intro}A, this treats the teacher as a static filter. Although the teacher can often produce valid solutions for simpler problems, it may fail to independently solve complex ``corner-case'' problems under a limited sampling budget. These difficult problems are then labeled as unsolved and discarded, even though they are precisely the samples from which the student could benefit most. This creates a data-efficiency bottleneck: the student is mainly trained on easy-to-medium examples, forming an artificial ``Teacher Ceiling'' that limits exposure to the hardest part of the problem distribution. Our empirical probing with Qwen3-32B on AIME 2025 shows that even with $N=64$ samples, the teacher still fails to generate any correct trajectory for approximately 13\% of hard problems.\footnote{This empirical observation was conducted using Qwen3-32B as the teacher model on AIME 2025.} This suggests that simply increasing the sampling budget may be inefficient for recovering such hard cases.

We argue that a teacher's failure to solve a problem independently does not necessarily imply a lack of latent reasoning capability. In many cases, the model may only need a small ``nudge'' to enter the correct search space. As illustrated in Figure~\ref{fig:intro}B, instead of discarding failed hard problems, we introduce hindsight hints, symbolized by the golden key. By conditioning generation on the ground-truth answer or intermediate guidance, the teacher can synthesize answer-consistent reasoning trajectories for problems that were previously rejected. This transforms discarded failures into additional training signals and enriches the student’s exposure to challenging reasoning patterns.

To address the ``Teacher Ceiling'' problem, we propose \textbf{Hindsight Entropy-Assisted Learning} (HEAL), an RL-free distillation framework inspired by a pedagogical process in which a teacher refines self-study notes into teachable lesson plans. HEAL consists of three components. First, \textbf{Guided Entropy-Assisted Repair} (GEAR) mimics an expert teacher’s self-repair process: the teacher first attempts the problem independently and consults hints only near the point of uncertainty. This is motivated by Vygotsky’s Zone of Proximal Development (ZPD) \cite{vygotsky1978mind} and scaffolding theory \cite{wood1976role}, where effective learning occurs between unaided and guided capability. In HEAL, entropy gradients are used to locate high-uncertainty breakpoints, where local hindsight hints provide scaffolding while retaining the teacher’s original reasoning prefix.

Second, a good teacher does not directly hand raw scratchpads to students, but refines them into structured lesson plans. We therefore introduce \textbf{Perplexity-Uncertainty Ratio Estimator} (PURE), a ratio-based filtering protocol for pedagogical quality control. Since hindsight-conditioned generation may produce shortcut-like rationales that force-fit reasoning to a known answer, PURE compares step-wise perplexity with answer uncertainty to identify abrupt and incoherent transitions. This filtering step reduces high-anomaly trajectories and encourages clearer step-by-step derivations.

Finally, knowledge transfer is organized through \textbf{Progressive Answer-guided Curriculum Evolution} (PACE), a three-stage curriculum distillation strategy. The student first learns from standard correct trajectories, then incorporates globally hinted solutions, and finally studies locally repaired trajectories from difficult corner cases. This educational progression builds a stable foundation before exposing the student to noisier but more informative hard examples.

Our main contributions are as follows:
\begin{enumerate}
    \item We propose HEAL, an interventionist distillation framework that uses hindsight-guided repair to recover additional training trajectories from failed teacher attempts.
    \item We design GEAR and PURE to respectively localize high-uncertainty repair points and filter shortcut-like answer-conditioned trajectories for higher-quality distillation data.
    \item We introduce PACE, a three-stage curriculum strategy, and conduct extensive experiments across multiple datasets and student models. Results show that HEAL consistently improves reasoning distillation over stronger rejection-sampling and curriculum baselines.
\end{enumerate}

\section{Related Work}

\textbf{Large Reasoning Model Distillation.}
Knowledge distillation provides an efficient alternative to reinforcement learning by transferring reasoning capabilities from strong teacher models to smaller students through supervised fine-tuning \cite{shridhar2023distilling}. Representative efforts, including the DeepSeek-R1-Distill series \cite{guo2025deepseek} and data-efficient methods such as s1 \cite{muennighoff2025s1} and LIMO \cite{Ye2025limo}, show that a small set of high-quality reasoning traces can substantially improve downstream performance. Recent studies further suggest that distillation can improve both accuracy and capability when teacher traces contain knowledge or reasoning patterns absent from the student \cite{kim2025reinforcement,yue2025does}, and may induce flexible reasoning behaviors that are difficult to obtain through Zero-RL alone on smaller models \cite{hu2025distillation}. However, standard rejection-sampling distillation is limited by the teacher's ability to independently produce correct trajectories, leaving hard problems underrepresented. HEAL addresses this bottleneck by using hindsight-guided local repair to recover additional answer-consistent trajectories.

\textbf{Knowledge-Guided Reasoning.}
Prior work has incorporated guidance prompts into reasoning model training to reduce exploration along erroneous paths \cite{li2025questa,liu2025ghpo,zhang2025edge}. These methods are mainly RL-based and often rely on richer process supervision, whereas our setting focuses on distillation using only final answers. Answer-conditioned rationale generation has also been explored in STaR \cite{Zelikman2022STaRBR} and RAVR \cite{lin2025ravr}: STaR bootstraps reasoning by rationalizing ground-truth answers, while RAVR uses prompting strategies to encourage more independent-looking solutions. However, global answer hints can be insufficient for hard mathematical corner cases and may produce shortcut rationales that drift toward the answer without a faithful derivation. HEAL differs by applying entropy-guided local repair and introducing PURE to filter high-anomaly, shortcut-like trajectories.

\textbf{Self-Distillation Paradigms.}
Recent self-distillation methods improve reasoning models through on-policy or iterative self-improvement within a single model family \cite{hubotter2026reinforcement,shenfeld2026self,zhao2026self}. In contrast, HEAL targets cross-model distillation, where a stronger teacher provides repairable trajectories for a smaller student. Its core components, including entropy-guided breakpoint selection and ratio-based shortcut filtering, are model-agnostic and could complement future self-distillation pipelines.

\section{Problem Formulation}

We formulate the reasoning distillation task as transferring the joint distribution of reasoning paths $p$ and responses $a$ from a large teacher model $\pi_T$ to a compact student model $\pi_S$, parameterized by $\theta_S$.

\vspace{0.5em}
\noindent \textbf{Data Elicitation.} 
We construct a distillation dataset $\mathcal{D}_{\text{RD}}$ via rejection sampling that approximates the teacher's reasoning distribution. Given a question set $\mathcal{Q}$, we sample rationale-response pairs from $\pi_T$ subject to a quality verification function $\Phi$:

\begin{equation}
\begin{split}
    \mathcal{D}_{\text{RD}} = \Big\{ (q, p, a) \mid (p, a) \sim \pi_T(\cdot|q), \\
    \Phi(q, p, a)=1, \ q \in \mathcal{Q} \Big\}
\end{split}
\label{eq:dataset_construction}
\end{equation}

Here, $\mathcal{D}_{\text{RD}}$ serves as the empirical ground truth, explicitly linking the teacher's latent reasoning capabilities to the training data available for the student.

\vspace{0.5em}
\noindent \textbf{Optimization Objective.} 
The goal is to optimize $\theta_S$ to minimize the expected negative log-likelihood over $\mathcal{D}_{\text{RD}}$. By applying the chain rule of probability, we decompose the objective into \textit{rationale modeling} and \textit{response generation}:

\begin{equation}
\begin{split}
    \mathcal{L}(\theta_S) = \mathbb{E}_{\mathcal{D}_{\text{RD}}} \Bigg[ 
    &- \underbrace{\sum_{t=1}^{|p|} \log \pi_S(p_t | q, p_{<t})}_{\mathcal{L}_{\text{reasoning}}} \\
    &- \underbrace{\sum_{k=1}^{|a|} \log \pi_S(a_k | q, p, a_{<k})}_{\mathcal{L}_{\text{response}}} \Bigg]
\end{split}
\label{eq:optimization_objective}
\end{equation}

This objective strictly compels $\pi_S$ to mimic the step-by-step reasoning process observed in $\mathcal{D}_{\text{RD}}$, rather than merely overfitting to the final answer $a$.

\section{Methodology}

\begin{figure*}[t]
\centering
\includegraphics[width=0.96\textwidth]{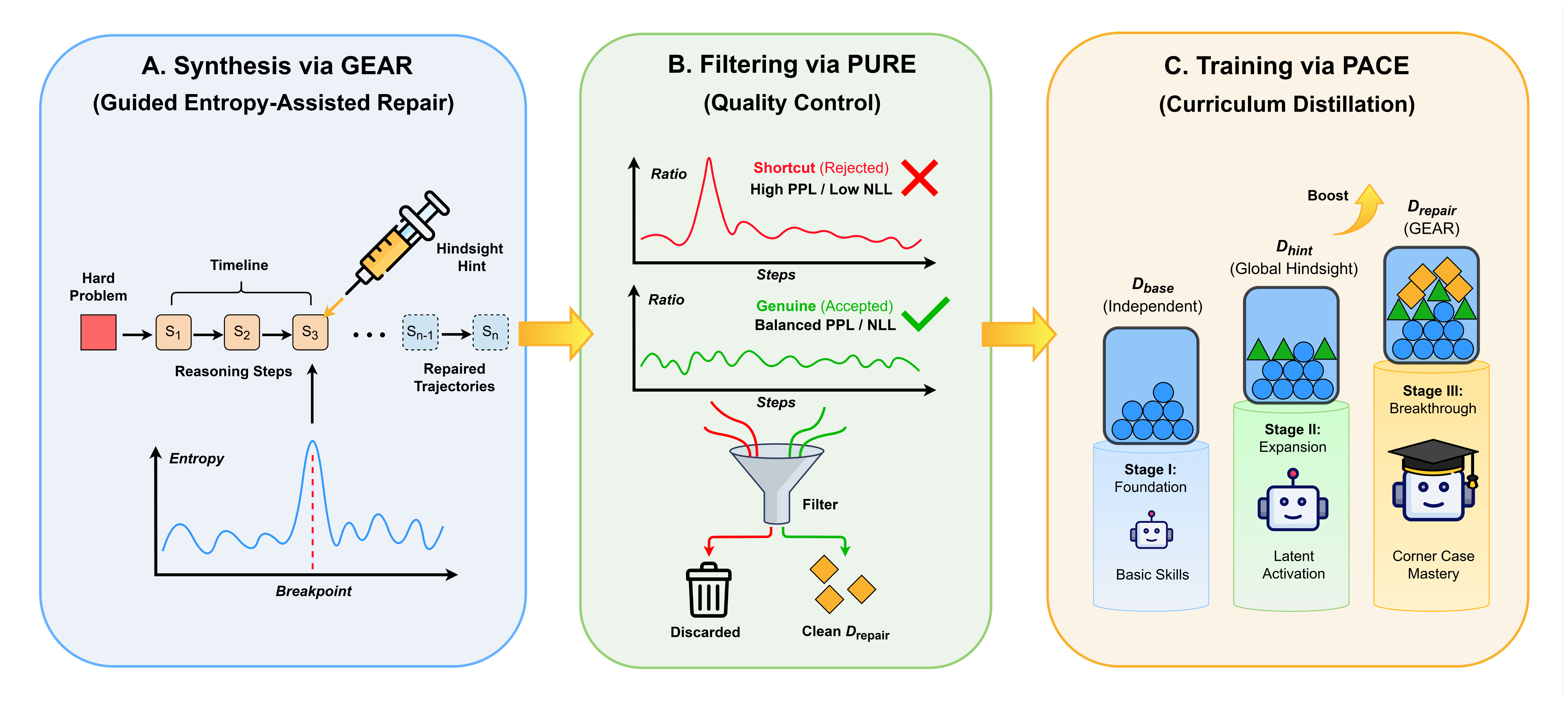}
\caption{The HEAL framework for reasoning distillation. } \label{fig:HEAL}
\end{figure*}


We propose HEAL, a unified framework designed to unlock the latent reasoning capabilities of teacher models for efficient distillation. As illustrated in Figure 2, HEAL synergizes three core modules to address the challenges of exploration, quality control, and knowledge transfer:

\begin{itemize}
    \item Synthesis via GEAR: An active intervention module that detects reasoning dead-ends via entropy dynamics and repairs them with targeted hindsight hints \footnote{Please refer to  Appendix A (Figure~\ref{fig:trajectory_guided}) for the prompt.}, synthesizing valid trajectories for previously intractable problems.
    \item Filtering via PURE: A rigorous quality control metric that decouples genuine reasoning breakthroughs from spurious shortcuts, ensuring the logical integrity of the synthesized data.
    \item Training via PACE: A stagewise distillation strategy that organizes training data from foundational to complex, enabling the student model to robustly absorb the repaired reasoning paths.
\end{itemize}

The formulation and implementation of these three modules are elaborated in the following subsections.

\subsection{Synthesis via GEAR}

Standard rejection sampling is often prohibitively inefficient for "hard" instances, where the probability of autonomously sampling a correct trajectory is vanishingly small. The goal of GEAR is to increase the yield of valid reasoning paths on these challenging instances by actively intervening in the teacher model's reasoning process at critical moments of uncertainty.

\paragraph{Theoretical Grounding.}
Our approach is grounded in the educational theory of the ZPD \cite{vygotsky1978mind}. We posit that for hard problems, the correct reasoning path lies within the teacher model's ZPD—it cannot be traversed independently but is accessible with appropriate scaffolding \cite{wood1976role}. GEAR operationalizes this by identifying the exact boundary of the ZPD as an estimated high-uncertainty breakpoint (where the teacher's reasoning confidence wavers) and providing a temporary scaffold (hindsight hint) to bridge the cognitive gap.

\paragraph{Identifying Critical Reasoning Breakpoint.}
We identify the precise boundary of independent capability, a likely transition point where uncertainty increases, by analyzing the entropy dynamics within the foundational phase of reasoning. Rather than reacting to minor fluctuations, we seek maximal cognitive dissonance, the largest entropy increase. For the $t$-th reasoning step $s_t=(w_{t,1},\ldots,w_{t,n_t})$, its token-averaged predictive entropy under teacher forcing is
\begin{equation}
\label{eq:step_entropy}
\begin{aligned}
\mathcal{H}(s_t)&=-\frac{1}{n_t}\sum_{j=1}^{n_t}\sum_{v\in\mathcal{V}}
p_{t,j}(v)\log p_{t,j}(v),\\
p_{t,j}(v)&=\pi_T\!\left(v\mid q,s_{<t},w_{t,<j}\right),
\end{aligned}
\end{equation}
where $\mathcal{V}$ is the teacher vocabulary. Formally, the Critical Reasoning Breakpoint $t^*$ is defined as the step exhibiting the maximum entropy surge within the initial segment of the trajectory:

\begin{equation}
t^* = \operatorname*{arg\,max}_{t} \{ \nabla \mathcal{H}(s_t) \mid 1 < t < \frac{L}{3} \} - 1 
\end{equation}

where
   $\nabla \mathcal{H}(s_t) = \mathcal{H}(s_t) - \mathcal{H}(s_{t-1})$ denotes the sudden increase in uncertainty. The search scope is constrained to the first third of the sequence ($t < L/3$).  This constraint is theoretically grounded in autoregressive generation dynamics, specifically aiming to suppress Error Cascading via Early-stage Anchoring. In complex mathematical reasoning, semantic information at the beginning of a trajectory serves as a critical structural "anchor." A logical deviation in this early planning phase often causes an irreversible probability shift in the state space, rendering subsequent reasoning entirely invalid regardless of computational accuracy. By intervening early, GEAR precisely targets and repairs these fundamental structural formulation errors rather than mere late-stage calculation slips. We then backtrack one step ($-1$) from the peak to anchor the intervention at the last stable logical state before the divergence occurs.

This strategy ensures that GEAR intervenes exactly at the most fragile link near an early high-uncertainty step in the reasoning chain, providing scaffolding where the model struggles the most.

\subsection{Filtering via PURE}
\label{sec:prdf}

A critical challenge in answer-conditioned generation is the phenomenon of "logical shortcuts". When provided with the ground-truth answer, language models are prone to generating "cheating" rationales that are syntactically coherent but logically disconnected (e.g. ``\textit{Since the reference answer is 36, the Final Answer is 36}''), often jumping abruptly to the final conclusion without valid derivation. Such trajectories act as noise during distillation.

To rigorously sanitize the synthesized datasets, we introduce PURE to decouple genuine reasoning breakthroughs from spurious shortcuts.

PURE identifies shortcuts by examining the relationship between a step's perplexity and the model's confidence in the reference answer after that step. We define the Suspicion Ratio $R_t$ as:

\begin{equation}
R_t = \frac{\text{PPL}(s_t | q, s_{<t})}{\text{NLL}(a | q, s_{\le t}) + \epsilon}
\end{equation}

where
    $\text{PPL}(s_t | q, s_{<t})$ denotes the step-wise perplexity evaluated by the teacher model $\pi_T$, quantifying the surprisal of generating the current step given the preceding context. $\text{NLL}(a | q, s_{\le t})$ is the answer uncertainty, defined as negative log-likelihood (NLL) of the ground-truth answer $a$ given the current step $s_t$. $\epsilon$ is a smoothing term to prevent division by zero. 


To operationalize this metric, we process each repaired trajectory by first excluding the final two steps, as these typically contain the concluding statement (e.g. ``\textit{The answer is X}''), which naturally exhibits high information gain but is logically valid. For the remaining steps, we compute the Suspicion Ratio ${R_t}$ sequentially, generating a ratio curve for the entire reasoning path. We then identify all local peaks within this curve and designate the global maximum peak value as the \textbf{Trajectory Anomaly Score} ($\mathcal{A}_{traj} = \max (R_t)$). This single scalar serves as a proxy for the worst-case logical violation within a trajectory. Finally, to construct the clean $\mathcal{D}_{repair}$ dataset, we rank all synthesized trajectories by their anomaly scores and prune the top $\lambda $ (e.g., $\lambda =0.2$), effectively filtering out the most egregious instances of shortcut learning while preserving the valid reasoning trajectories.

\subsection{Training via PACE}

While GEAR and PURE provide high-quality reasoning trajectories for complex problems, directly exposing a student model to these advanced samples can lead to training instability or catastrophic forgetting of foundational skills. To mitigate this, we propose PACE, a three-stage distillation strategy designed to align the student's learning trajectory with the complexity of the data.

We categorize the training data into three tiers of complexity:
\begin{itemize}
    \item $\mathcal{D}_{base}$ (Independent Reasoning): Standard trajectories obtained via rejection sampling where the teacher solves problems without assistance.
    \item $\mathcal{D}_{hint}$ (Global Hindsight): Trajectories generated by providing the ground-truth answer as a global hint in the prompt \footnote{Please refer to  Appendix A (Figure~\ref{fig:answer_guided})  for the prompt.}, filtered by PURE. These represent problems within the teacher's latent capability but requiring goal-oriented guidance.
    \item $\mathcal{D}_{repair}$ (Local Repair): Trajectories synthesized via GEAR's entropy-guided intervention and filtered by PURE. These represent the "hardest" corner cases where the teacher faces reasoning dead-ends even with global hints.
\end{itemize}

PACE organizes the distillation process into three progressive stages:

\textbf{Stage I: Foundation Alignment.}
The student is trained exclusively on $\mathcal{D}_{base}$.

\begin{equation}
\mathcal{L}_{I} = \mathcal{L}_{SFT}(\theta, \mathcal{D}_{base}) 
\end{equation}

This phase ensures the student masters fundamental arithmetic rules and the teacher's baseline reasoning patterns, establishing a stable instruction-following capability.

\textbf{Stage II: Latent Expansion.}
We introduce $\mathcal{D}_{hint}$ to unlock reasoning paths that are valid but require answer-conditioned navigation. To prevent catastrophic forgetting, we mix these with the foundational data (e.g., $\mathcal{D}_{base} \cup \mathcal{D}_{hint}$).

\begin{equation}
\mathcal{L}_{II} = \mathcal{L}_{SFT}(\theta, \mathcal{D}_{base} + \mathcal{D}_{hint}) 
\end{equation}

This phase bridges the gap between independent solving and complex repairs, expanding the student's solution space to medium-difficulty problems.

\textbf{Stage III: Hard Case Adaptation.} 
The final phase targets the most intractable problems. We integrate the GEAR-synthesized trajectories $\mathcal{D}_{repair}$ into the mix, typically upsampling them to emphasize handling cognitive impasses.

\begin{equation}
\mathcal{L}_{III} = \mathcal{L}_{SFT}(\theta, \mathcal{D}_{base} + \mathcal{D}_{hint} + \mathcal{D}_{repair})
\end{equation}

By progressively exposing the student to problems with different difficulties, PACE enhances the student's proficiency in handling complex corner cases, thereby maximizing the distillation effect and narrowing the performance gap between the student and the teacher.

\begin{table*}[!ht]
  \centering
  \resizebox{1.0\textwidth}{!}{
  \begin{tabular}{>{\centering\arraybackslash}m{2.2cm}|c|ccccc|c}
  \toprule
  \toprule
  \textbf{Models} & \textbf{Method}
  & \textbf{AIME 2024} & \textbf{AIME 2025}
  & \textbf{MATH 500} & \textbf{OlympiadBench}
  & \textbf{GPQA} & \textbf{AVG.} \\
  \midrule

  \multirow{8}{*}{\textbf{\makecell{Qwen2.5-14B\\-Instruct}}}
  & Origin
  & 19.17\textsubscript{$\pm$1.49}
  & 20.00\textsubscript{$\pm$0.00}
  & 81.15\textsubscript{$\pm$0.18}
  & 42.44\textsubscript{$\pm$0.38}
  & 40.34\textsubscript{$\pm$0.32}
  & 40.62 \\

  & LIMO
  & 16.25\textsubscript{$\pm$1.67}
  & 13.75\textsubscript{$\pm$1.14}
  & 78.00\textsubscript{$\pm$0.44}
  & 35.15\textsubscript{$\pm$0.22}
  & 32.64\textsubscript{$\pm$0.26}
  & 35.16 \\

  & STaR
  & 24.58\textsubscript{$\pm$2.06}
  & 23.13\textsubscript{$\pm$0.83}
  & 83.45\textsubscript{$\pm$0.26}
  & 46.52\textsubscript{$\pm$0.46}
  & 40.66\textsubscript{$\pm$0.47}
  & 43.67 \\

  & RS@30-SFT
  & 32.92\textsubscript{$\pm$1.67}
  & 29.79\textsubscript{$\pm$1.48}
  & 87.05\textsubscript{$\pm$0.21}
  & 54.11\textsubscript{$\pm$0.30}
  & 42.80\textsubscript{$\pm$0.36}
  & 49.33 \\

  & RS@128-SFT
  & 35.63\textsubscript{$\pm$2.01}
  & 32.50\textsubscript{$\pm$1.49}
  & 87.13\textsubscript{$\pm$0.24}
  & 55.41\textsubscript{$\pm$0.42}
  & 43.18\textsubscript{$\pm$0.38}
  & 50.77 \\

  & RS@CM-SFT
  & 34.17\textsubscript{$\pm$1.92}
  & 34.79\textsubscript{$\pm$2.10}
  & 87.68\textsubscript{$\pm$0.26}
  & 56.50\textsubscript{$\pm$0.38}
  & 42.49\textsubscript{$\pm$0.32}
  & 51.12 \\

  & Curriculum-SFT
  & 36.46\textsubscript{$\pm$0.83}
  & 33.13\textsubscript{$\pm$1.48}
  & 87.83\textsubscript{$\pm$0.20}
  & 57.54\textsubscript{$\pm$0.44}
  & 43.69\textsubscript{$\pm$0.27}
  & 51.73 \\

  & \textbf{HEAL}
  & \textbf{53.96}\textsubscript{$\pm$1.81}
  & \textbf{40.42}\textsubscript{$\pm$1.67}
  & \textbf{90.63}\textsubscript{$\pm$0.23}
  & \textbf{62.28}\textsubscript{$\pm$0.36}
  & \textbf{47.22}\textsubscript{$\pm$0.38}
  & \textbf{58.90} \\

  \midrule

  \multirow{8}{*}{\textbf{\makecell{Qwen3-4B\\-Base}}}
  & Origin
  & 7.92\textsubscript{$\pm$2.06}
  & 4.17\textsubscript{$\pm$1.49}
  & 67.13\textsubscript{$\pm$0.34}
  & 38.17\textsubscript{$\pm$0.21}
  & 8.08\textsubscript{$\pm$0.27}
  & 25.09 \\

  & LIMO
  & 13.54\textsubscript{$\pm$0.83}
  & 22.50\textsubscript{$\pm$1.92}
  & 77.68\textsubscript{$\pm$0.28}
  & 38.43\textsubscript{$\pm$0.34}
  & 12.18\textsubscript{$\pm$0.32}
  & 32.87 \\

  & STaR
  & 11.88\textsubscript{$\pm$2.10}
  & 8.96\textsubscript{$\pm$1.60}
  & 72.33\textsubscript{$\pm$0.43}
  & 38.93\textsubscript{$\pm$0.39}
  & 9.47\textsubscript{$\pm$0.36}
  & 28.31 \\

  & RS@30-SFT
  & 19.58\textsubscript{$\pm$1.14}
  & 18.96\textsubscript{$\pm$2.01}
  & 83.25\textsubscript{$\pm$0.21}
  & 39.83\textsubscript{$\pm$0.18}
  & 14.14\textsubscript{$\pm$0.27}
  & 35.15 \\

  & RS@128-SFT
  & 20.63\textsubscript{$\pm$1.81}
  & 21.46\textsubscript{$\pm$2.10}
  & 84.48\textsubscript{$\pm$0.30}
  & 39.41\textsubscript{$\pm$0.35}
  & 16.22\textsubscript{$\pm$0.32}
  & 36.44 \\

  & RS@CM-SFT
  & 22.29\textsubscript{$\pm$2.01}
  & 19.79\textsubscript{$\pm$0.83}
  & 84.25\textsubscript{$\pm$0.28}
  & 40.50\textsubscript{$\pm$0.33}
  & 15.53\textsubscript{$\pm$0.36}
  & 36.47 \\

  & Curriculum-SFT
  & 21.67\textsubscript{$\pm$1.72}
  & 20.83\textsubscript{$\pm$1.49}
  & 84.53\textsubscript{$\pm$0.32}
  & 39.93\textsubscript{$\pm$0.40}
  & 17.11\textsubscript{$\pm$0.32}
  & 36.81 \\

  & \textbf{HEAL}
  & \textbf{30.21}\textsubscript{$\pm$0.83}
  & \textbf{33.33}\textsubscript{$\pm$0.00}
  & \textbf{87.18}\textsubscript{$\pm$0.25}
  & \textbf{42.17}\textsubscript{$\pm$0.32}
  & \textbf{20.14}\textsubscript{$\pm$0.32}
  & \textbf{42.60} \\

  \midrule

  \multirow{8}{*}{\textbf{\makecell{gemma-4\\-E4B-it}}}
  & Origin
  & 59.38\textsubscript{$\pm$1.81}
  & 39.58\textsubscript{$\pm$1.14}
  & 81.65\textsubscript{$\pm$0.18}
  & 67.67\textsubscript{$\pm$0.22}
  & 56.06\textsubscript{$\pm$0.47}
  & 60.87 \\

  & LIMO
  & 45.83\textsubscript{$\pm$1.92}
  & 32.29\textsubscript{$\pm$2.01}
  & 72.33\textsubscript{$\pm$0.38}
  & 60.26\textsubscript{$\pm$0.49}
  & 47.41\textsubscript{$\pm$0.50}
  & 51.62 \\

  & STaR
  & 61.04\textsubscript{$\pm$1.60}
  & 41.46\textsubscript{$\pm$2.10}
  & 82.55\textsubscript{$\pm$0.46}
  & 68.17\textsubscript{$\pm$0.41}
  & 56.69\textsubscript{$\pm$0.45}
  & 61.98 \\

  & RS@30-SFT
  & 62.92\textsubscript{$\pm$1.67}
  & 42.50\textsubscript{$\pm$1.49}
  & 83.58\textsubscript{$\pm$0.31}
  & 68.30\textsubscript{$\pm$0.39}
  & 57.95\textsubscript{$\pm$0.59}
  & 63.05 \\

  & RS@128-SFT
  & 65.63\textsubscript{$\pm$2.01}
  & 43.13\textsubscript{$\pm$0.83}
  & 84.15\textsubscript{$\pm$0.35}
  & 67.41\textsubscript{$\pm$0.43}
  & 58.21\textsubscript{$\pm$0.52}
  & 63.70 \\

  & RS@CM-SFT
  & 64.17\textsubscript{$\pm$1.92}
  & 43.96\textsubscript{$\pm$1.81}
  & 84.28\textsubscript{$\pm$0.34}
  & 68.20\textsubscript{$\pm$0.38}
  & 57.70\textsubscript{$\pm$0.45}
  & 63.66 \\

  & Curriculum-SFT
  & 66.25\textsubscript{$\pm$1.14}
  & 43.75\textsubscript{$\pm$2.06}
  & 84.38\textsubscript{$\pm$0.38}
  & 68.31\textsubscript{$\pm$0.49}
  & 58.59\textsubscript{$\pm$0.47}
  & 64.26 \\

  & \textbf{HEAL}
  & \textbf{70.83}\textsubscript{$\pm$1.49}
  & \textbf{46.04}\textsubscript{$\pm$1.34}
  & \textbf{86.23}\textsubscript{$\pm$0.31}
  & \textbf{68.96}\textsubscript{$\pm$0.30}
  & \textbf{60.35}\textsubscript{$\pm$0.47}
  & \textbf{66.48} \\

  \bottomrule
  \bottomrule
  \end{tabular}
  }
\caption{Performance comparison of models across benchmarks. Results are reported as mean $\pm$ standard deviation. Best results within each model group are highlighted in bold.}
\label{tab:main_performance}
\end{table*}

\section{Experiments}

\subsection{Experimental Settings}


\subsubsection{Benchmarks}
\label{sec:appendix_Benchmarks}
We evaluate models on five widely used and challenging reasoning benchmarks: 
\begin{itemize}
    \item \textbf{MATH 500} \citep{hendrycksmath2021, lightman2023let}: It is a representative subset of the MATH dataset, comprising 500 problems that span diverse mathematical disciplines and difficulty levels.
    \item \textbf{AIME 2024 \& AIME 2025} \citep{AIME24, AIME25}: Each consists of 30 problems from the American Invitational Mathematics Examination.
    \item \textbf{OlympiadBench} \citep{he2024olympiadbench}: It is a comprehensive and rigorous benchmark featuring elite-level problems from international Olympiads. We evaluate on the subset of 675 English math problems.
    \item \textbf{GPQA} \citep{rein2024gpqa}: A highly challenging benchmark spanning biology, chemistry, and physics formulated by domain experts. We evaluate on the Diamond subset to rigorously test expert-level scientific reasoning.
\end{itemize}

\subsubsection{Teacher Model and Student Models}

For reasoning distillation, we employ Qwen3-32B\footnote{\url{https://huggingface.co/Qwen/Qwen3-32B}} as the teacher model to generate high-quality reasoning trajectories. For the student models, we utilize Qwen2.5-14B-Instruct\footnote{\url{https://huggingface.co/Qwen/Qwen2.5-14B-Instruct}}, Qwen3-4B-Base\footnote{\url{https://huggingface.co/Qwen/Qwen3-4B-Base}}, and gemma-4-E4B-it\footnote{\url{https://huggingface.co/google/gemma-4-E4B-it}}. This selection spans multiple model families and parameter scales and includes both base and instruction-tuned checkpoints.

\subsubsection{Baseline Methods}
We compare our proposed approach with six representative baselines:
\begin{itemize}
    \item \textbf{RS@30-SFT} \cite{ouyang2022training}: We sample 30 independent trajectories per training problem from the teacher model and retain only those with correct final answers, and fine-tune the student on them. This corresponds to the standard rejection-sampling SFT baseline.

    \item \textbf{LIMO} \citep{Ye2025limo}: LIMO employs hand-crafted rules to select the reasoning paths sampled from the teacher model. Following this methodology, we reproduced a comparable data selection pipeline for our training dataset. 

    \item \textbf{STaR} \citep{Zelikman2022STaRBR}: STaR boosts reasoning by iteratively fine-tuning on self-generated correct rationales.

    \item \textbf{RS@128-SFT}: Rejection sampling with 128 attempts per problem to ensure sufficient sampling coverage, where only trajectories with correct answers are retained for SFT.

    \item \textbf{RS@CM-SFT}: A cost-matched rejection sampling baseline that adopts the same sampling budget as HEAL. Problems for which fewer than 15 of the 30 sampled trajectories yield correct final answers are labeled as hard ($H$); among these, problems with at most one trajectory yielding a correct final answer are labeled as extremely hard ($E \subseteq H$). For each problem $q_i$, the sampling budget is $K_i^{CM} = 30 + 30 \cdot 1[q_i \in H] + 200 \cdot 1[q_i \in E]$, where normal, hard, and extremely hard problems thus receive 30, 60, and 260 teacher generations, respectively. Unlike HEAL, this baseline performs standard rejection sampling without answer-conditioned local repair and retains only trajectories with correct final answers.
    

    \item \textbf{Curriculum-SFT} \citep{wen2025csft}: This method aims to enhance Long-CoT reasoning capabilities through a multi-stage training process characterized by progressive difficulty.

\end{itemize}



\subsubsection{Training Details}

We implement the training pipeline via the MS-SWIFT\footnote{\url{https://github.com/modelscope/ms-swift}} framework using 919 problems from the AIME\footnote{\url{https://huggingface.co/datasets/gneubig/aime-1983-2024}. We removed all AIME 2024 problems from the training set.} spanning 1985 to 2023. To construct the distillation  dataset, we employ a dynamic sampling strategy with a temperature of 0.7 and top\_p of 0.8. The data elicitation process proceeds in three stages based on problem difficulty:

1. Initial Sampling ($\mathcal{D}_{base}$): We first sample 30 reasoning trajectories per problem using the teacher model. Problems for which fewer than 15 of the 30 sampled trajectories yield correct final answers are identified as hard instances.

2. Global Hindsight Sampling ($\mathcal{D}_{hint}$): For these hard instances, we perform an additional round of sampling (30 trajectories per problem) using answer-conditioned prompts to elicit valid paths.

3. Local Repair Sampling ($\mathcal{D}_{repair}$): For "extremely hard" problems—defined as those yielding $\le 1$ correct trajectory in the initial phase—we select 10 incorrect paths and apply GEAR to repair them, generating 20 candidate repaired trajectories for each error path.

All answer-conditioned trajectories in $\mathcal{D}_{hint}$ and $\mathcal{D}_{repair}$ undergo PURE filtering. The final distilled dataset consists of 5,777 samples for $\mathcal{D}_{base}$, 1,108 samples for $\mathcal{D}_{hint}$, and 2,544 samples for $\mathcal{D}_{repair}$. In the implementation of PURE, we segment the reasoning trajectory into steps using "\texttt{\textbackslash n\textbackslash n}" as the delimiter. $\lambda$ is set to 0.2, and the smoothing term $\epsilon$ is set to 0.01. Our training was conducted over 5 epochs with a learning rate of $1\times10^{-5}$ and a global batch size of 8. Experiments were executed on 8$\times$NVIDIA H100 GPUs, leveraging DeepSpeed ZeRO-3 \citep{rasley2020deepspeed,rajbhandari2020zero} and FlashAttention-2 \citep{dao2023flashattention} for efficiency.

\subsubsection{Evaluation Details}
For each trained checkpoint, we sample N independent completions per problem (N = 16 for AIME 2024 and AIME 2025; N=8 for MATH 500, OlympiadBench, and GPQA) and report the average single-sample accuracy over all completions. The reported standard deviation is computed over repeated decoding runs.

To strictly assess reasoning integrity, final answers are extracted from the generated reasoning chains and verified against the ground truth via exact match or symbolic equivalence. Regarding inference configuration, we set the maximum generation length to 32,768 tokens per completion. AIME 2024, AIME 2025, MATH 500, and GPQA were evaluated using EvalScope\footnote{\url{https://github.com/modelscope/evalscope}}, whereas OlympiadBench was assessed via the Qwen2.5-Math\footnote{\url{https://github.com/QwenLM/Qwen2.5-Math}} framework. For gemma-4-E4B-it, we performed dedicated code adaptations to ensure compatibility with all five evaluation benchmarks.

\section{Results and Analysis}
\subsection{Main Results}

Table~\ref{tab:main_performance} shows that HEAL consistently outperforms all baselines across the three student models. On Qwen2.5-14B-Instruct, HEAL achieves an average accuracy of 58.90\%, surpassing RS@30-SFT, RS@CM-SFT, and RS@128-SFT by 9.57, 7.78, and 8.13 percentage points, respectively. These gains indicate that HEAL improves reasoning distillation beyond simply increasing the rejection-sampling budget. The improvement is most pronounced on challenging math benchmarks: HEAL reaches 53.96\% on AIME 2024, outperforming the strongest baseline, Curriculum-SFT, by 17.50 percentage points, while also maintaining a consistent lead on OlympiadBench. The gains on GPQA further provide preliminary evidence of transfer beyond mathematics.

Compared with specialized distillation methods such as LIMO and self-bootstrapping methods such as STaR, HEAL yields more stable improvements across different model families and base/instruction-tuned student models. We attribute this robustness to its combination of targeted repair, shortcut filtering, and curriculum-based training, which together extract additional training signals from hard problems while reducing reliance on low-quality answer-conditioned trajectories.

\subsection{Ablation Study}


        

  \begin{table}[!ht]
      \centering
      \resizebox{\linewidth}{!}{
          \begin{tabular}{lccccc}
          \toprule
          \textbf{Method}
          & \textbf{AIME 24}
          & \textbf{AIME 25}
          & \textbf{MATH 500}
          & \textbf{OlympiadBench}
          & \textbf{GPQA} \\
          \midrule

          \textbf{HEAL}
          & \textbf{53.96}
          & \textbf{40.42}
          & \textbf{90.63}
          & \textbf{62.28}
          & \textbf{47.22} \\

          \midrule

          \quad Random Breakpoint
          & 50.21
          & 36.46
          & 89.60
          & 57.93
          & 46.28 \\

          \quad w/o PACE
          & 43.33
          & 36.46
          & 88.68
          & 58.48
          & 46.40 \\

          \quad w/o PURE
          & 42.08
          & 36.46
          & 88.45
          & 58.13
          & 45.64 \\

          \quad w/o GEAR
          & 42.29
          & 35.42
          & 89.58
          & 58.35
          & 46.34 \\

          \bottomrule
          \end{tabular}
      }
    \caption{The ablation study results of HEAL and its variants on Qwen2.5-14B-Instruct.}
    \vspace{-0.3cm}
    \label{tab:key-component}
  \end{table}

To rigorously validate the individual contributions of each component within the HEAL framework, we conducted a component-wise ablation study on Qwen2.5-14B-Instruct. Table \ref{tab:key-component} summarizes the performance across five benchmarks.

\paragraph{Impact of Entropy-Guided Synthesis (GEAR).}
We first analyze the synthesis strategy. Replacing the entropy-guided breakpoint detection with a Random Breakpoint strategy results in a noticeable performance drop (e.g., from 53.96\% to 50.21\% on AIME 2024). This suggests that reasoning failures are better handled by uncertainty-aware localization than by random intervention; utilizing entropy dynamics to pinpoint the exact moment of cognitive dissonance is crucial for effective repair. Furthermore, removing the local repair mechanism (w/o GEAR) and relying solely on global answer hints reduces accuracy on AIME 2024 from 53.96\% to 42.29\%, an absolute decrease of 11.67 percentage points. This indicates that global guidance alone is less effective than localized repair for the hard cases considered here; fine-grained, step-wise repairs are essential to bridge the reasoning gap.

\paragraph{Impact of Quality Control (PURE).}
Removing the filtering protocol (w/o PURE) results in the most severe performance degradation alongside w/o GEAR, plummeting AIME 2024 accuracy to 42.08\%. This highlights the detrimental impact of "shortcut learning". Without PURE, the training data likely contains more shortcut-like trajectories, which may hurt downstream performance. The significant gap between HEAL and w/o PURE underscores that the quality of hindsight data is as critical as its quantity. To verify that PURE improves trajectory quality rather than merely reducing data volume, we conduct a GPT-5.4-based audit in Appendix~\ref{app:trajectory_quality}. The audit shows fewer invalid or shortcut-based rationales after PURE filtering, especially in $D_{\text{repair}}$, supporting our observation that w/o PURE suffers from lower-quality answer-conditioned trajectories.

\paragraph{Impact of Curriculum Training (PACE).}
Finally, discarding the three-stage curriculum in favor of monolithic data mixing (w/o PACE) causes a sharp decline in performance (43.33\% on AIME 24). Despite having access to the same high-quality data, direct mixing fails to stabilize the learning process. This demonstrates that PACE is pivotal: it allows the student to consolidate foundational capabilities before tackling the high-complexity, entropy-repaired trajectories, thereby preventing catastrophic forgetting and ensuring robust knowledge transfer.

In summary, the synergy of all three modules—GEAR, PURE, and PACE—is required to achieve state-of-the-art distillation performance.


\paragraph{Impact of Truncation Percentage Threshold }
\label{sec:appendix_a}



\begin{figure}[!ht]
\centering
\includegraphics[width=0.9\linewidth]
{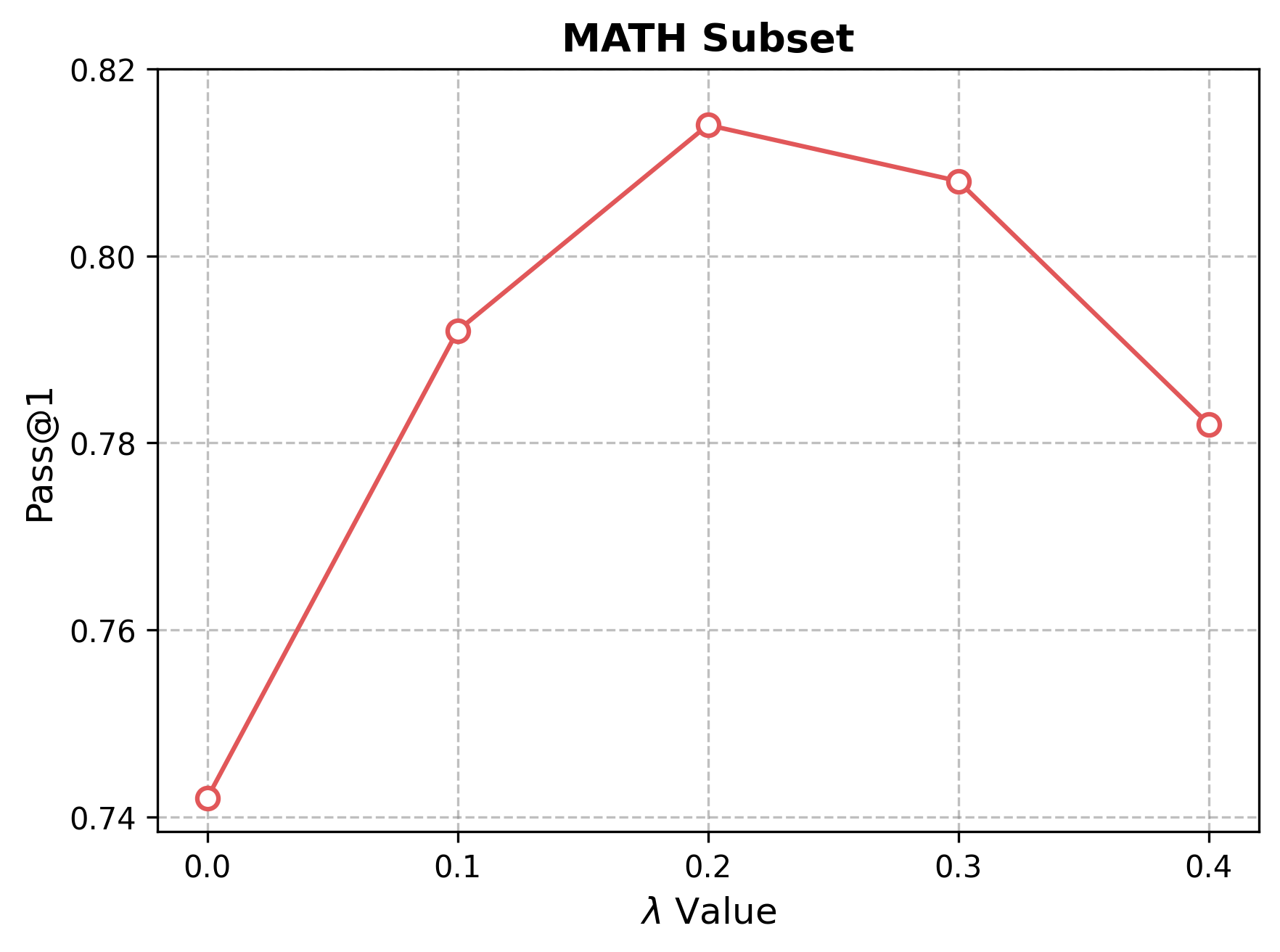}
\caption{Sensitivity analysis of $\lambda$ values on the MATH subset.} \label{fig:math_result}
\end{figure}

To investigate the impact of the truncation percentage threshold $\lambda$ for PURE, we conducted grid search validation on a subset of the MATH dataset using Qwen2.5-14B-Instruct. Specifically, we randomly sampled 500 problems with difficulty level 5 from the MATH training set to create this subset. As illustrated in Figure~\ref{fig:math_result}, evaluation results across different $\lambda$ values demonstrate that the model achieves optimal performance when $\lambda=0.2$. We attribute this to the delicate trade-off between data purity and diversity. When $\lambda$ is too low, the filter is overly permissive, allowing a significant fraction of ``shortcut'' trajectories that degrade performance. Conversely, an overly aggressive threshold risks discarding valid high-perplexity steps that represent genuine cognitive breakthroughs, thereby reducing the diversity of reasoning patterns available for distillation.

\paragraph{Impact of Search Scope in GEAR}
\label{sec:appendix_c}

\begin{figure}[!ht]
\centering
\includegraphics[width=0.88\linewidth]{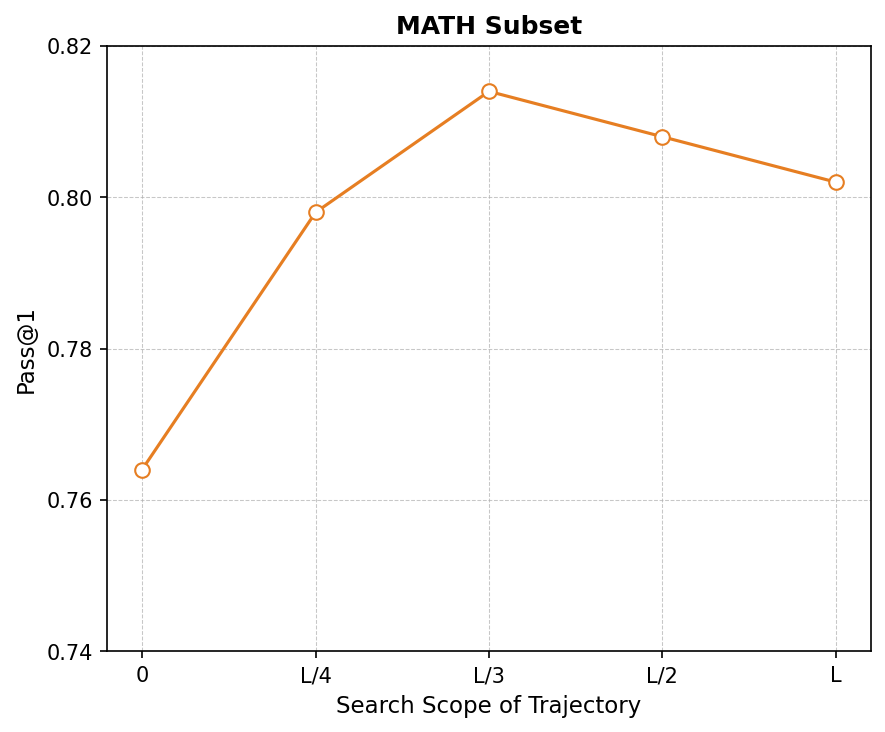}
\caption{Sensitivity analysis of search scope of trajectory on the MATH subset.} \label{fig:math_bench_gear}
\end{figure}

To empirically validate our decision to constrain the entropy-increase search scope to the first third of the sequence, we conduct a grid search on a subset of the MATH dataset using Qwen2.5-14B-Instruct. Specifically, we randomly sampled 500 problems with difficulty level 5 from the MATH training set to create this evaluation subset. We vary the scope threshold $\alpha$ from $0$ to $L$. As illustrated in Figure~\ref{fig:math_bench_gear}, performance peaks precisely at $\alpha=L/3$ and degrades monotonically as the boundary expands. This validates our theoretical premise for Early-stage Anchoring. If the scope is excessively narrow (e.g., $\alpha \le L/4$), GEAR fails to capture the structural formulation errors---often signaled by prominent early entropy surges---that act as critical anchors during the early planning phase, allowing Error Cascading to propagate unchecked. Conversely, an overly broad scope (e.g., $\alpha \ge L/2$) dilutes the entropy detection signal with late-stage calculation slips and extraneous steps, masking the root structural errors. Ultimately, the $L/3$ threshold establishes an optimal balance, effectively encompassing the critical structural window while filtering out the entropy noise of subsequent reasoning stages.

\section{Conclusion}


In this work, we introduced HEAL, an RL-free reasoning distillation framework that augments standard rejection sampling with hindsight-guided local repair, shortcut filtering, and curriculum training. Experiments across three student models and five benchmarks show that HEAL outperforms stronger rejection-sampling and curriculum baselines. Future work may combine HEAL with RL or stronger step-level verifiers to further improve trajectory faithfulness.

\section*{Limitations}
Despite its effectiveness, HEAL has three primary limitations. First, our approach assumes the target problem lies within the teacher model's ZPD. For problems that require domain knowledge or reasoning primitives completely absent from the teacher's pre-training distribution, even active intervention may fail to elicit valid reasoning paths. The second limitation is the additional computational overhead incurred by the step-wise calculation of Perplexity and NLL in the PURE module. However, we emphasize that this is strictly an offline cost and does not impose any latency during inference. Third, HEAL relies on answer-conditioned trajectory synthesis, which may still produce plausible but logically invalid rationalizations. Although PURE and our trajectory audit suggest that such problematic trajectories are reduced after filtering, PURE remains a heuristic rather than a formal step-level verifier. 

\section*{Ethics Statement}
This work does not involve human subjects, private user data, or any personally identifiable information. All datasets and model resources used in our experiments are publicly available and were used in accordance with their respective licenses and terms of use. During the writing process, AI-based tools were used only for grammar checking and language polishing.

\section*{Acknowledgments}
This work was supported by the National Natural Science Foundation of China Enterprise Innovation and Development Joint Fund Project (Grant No.~U24B20181).

\bibliography{acl_latex}

\clearpage 
\newpage
\appendix


\section{Teacher Sampling via Hindsight Prompting}
\label{sec:appendix}

\begin{figure}[!ht]

\centering
\includegraphics[width=0.47\textwidth]{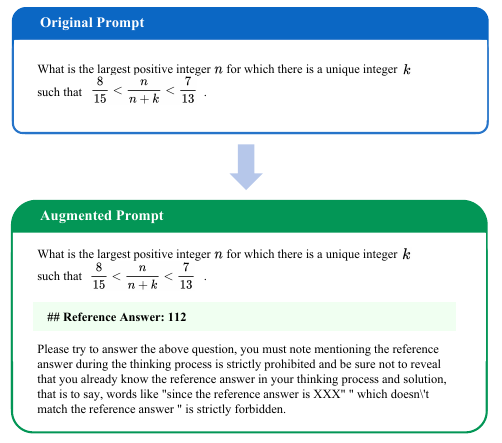}
\caption{Providing the ground-truth answer as a global hint in the prompt.} \label{fig:answer_guided}
\end{figure}

We utilize the prompt template in Figure \ref{fig:answer_guided} to sample high-quality trajectories from the teacher model, conditioning the generation on the 
ground-truth answer as a global hint. 

\begin{figure}[!ht]
\centering
\includegraphics[width=0.47\textwidth]{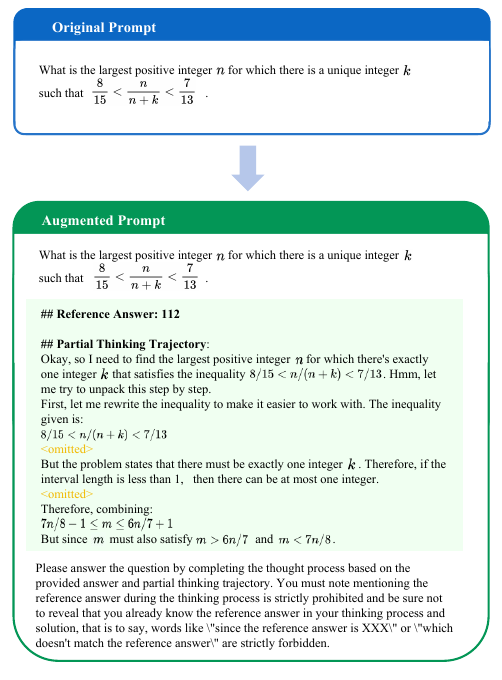}
\caption{Providing the ground-truth answer and partial trajectory as local repair guidance in the prompt.} \label{fig:trajectory_guided}
\end{figure}

For extremely challenging problems, we use the template in Figure \ref{fig:trajectory_guided}, which provides both the ground-truth answer and a GEAR-derived partial trajectory to elicit a repaired reasoning path.




\section{Trajectory Quality Validation}
\label{sec:appendix_e}

\label{app:trajectory_quality}

\begin{figure*}[!ht]
\centering
\includegraphics[width=1\textwidth]{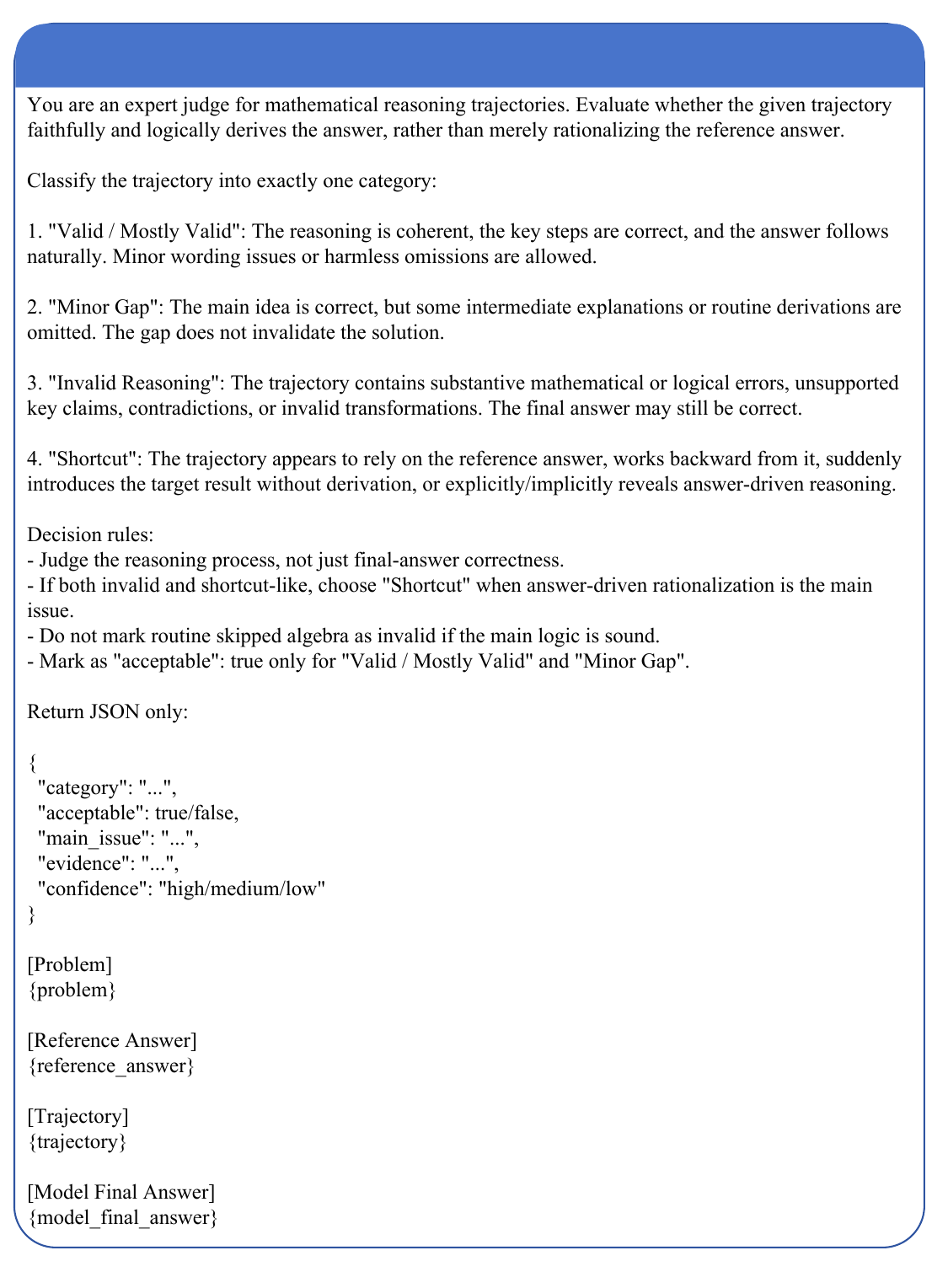}
\caption{Prompt template for GPT-5.4-based trajectory evaluation.} \label{fig:gpt5_prompt}
\end{figure*}

\begin{table*}[!ht]
\centering
\small
\setlength{\tabcolsep}{3pt}
\begin{tabular}{lcccccc}
\toprule
\textbf{Source} 
& \textbf{Valid / Mostly Valid} 
& \textbf{Minor Gap} 
& \textbf{Invalid Reasoning} 
& \textbf{Shortcut } 
& \textbf{Acceptable} 
& \textbf{Problematic} \\
\midrule
$D_{\text{base}}$ 
& 25 
& 4 
& 1 
& 0 
& 29 / 30 
& 1 / 30 \\
$D_{\text{hint}}$ before PURE 
& 21 
& 4 
& 3 
& 2 
& 25 / 30 
& 5 / 30 \\
$D_{\text{hint}}$ after PURE 
& 25 
& 3 
& 1 
& 1 
& 28 / 30 
& 2 / 30 \\
$D_{\text{repair}}$ before PURE 
& 17 
& 5 
& 4 
& 4 
& 22 / 30 
& 8 / 30 \\
$D_{\text{repair}}$ after PURE 
& 23 
& 4 
& 2 
& 1 
& 27 / 30 
& 3 / 30 \\
PURE-rejected trajectories 
& 8 
& 5 
& 7 
& 10 
& 13 / 30 
& 17 / 30 \\
\bottomrule
\end{tabular}
\caption{
Trajectory quality audit using GPT-5.4 as an LLM judge. Each row contains 30 randomly sampled trajectories. ``Acceptable'' denotes the sum of \textit{Valid / Mostly Valid} and \textit{Minor Gap}, while ``Problematic'' denotes the sum of \textit{Invalid Reasoning} and \textit{Shortcut}. The results suggest that PURE reduces the fraction of problematic answer-conditioned trajectories and that PURE-rejected samples contain substantially more invalid or shortcut rationales.
}
\label{tab:trajectory_quality_audit}
\end{table*}

A central concern of answer-conditioned reasoning distillation is that the teacher may generate rationales that are merely consistent with the provided answer, rather than logically valid. This issue is particularly relevant for \textsc{HEAL}, since both hindsight-guided trajectories and repaired trajectories are generated with access to the reference answer. To examine whether our filtering pipeline improves the quality of the synthesized reasoning trajectories, we conduct an additional trajectory-level audit using GPT-5.4 as an external LLM judge. The detailed prompt template employed for GPT-5.4-based trajectory evaluation is provided in Figure~\ref{fig:gpt5_prompt}.


\paragraph{Audit setup.}
We randomly sample 30 trajectories from each of the following sources: 
(i) $D_{\text{base}}$, the standard rejection-sampling trajectories generated without answer hints;
(ii) $D_{\text{hint}}$ before PURE filtering;
(iii) $D_{\text{hint}}$ after PURE filtering;
(iv) $D_{\text{repair}}$ before PURE filtering;
(v) $D_{\text{repair}}$ after PURE filtering; and
(vi) trajectories rejected by PURE. 
For each sampled trajectory, the judge is given the original problem, the final answer, and the full reasoning trajectory. The judge is instructed to evaluate only the logical validity and faithfulness of the reasoning process, rather than the final answer alone.

\paragraph{Annotation categories.}
Each trajectory is assigned to exactly one of four mutually exclusive categories:
\begin{itemize}
    \item \textbf{Valid / Mostly Valid}: the reasoning is logically coherent, the key derivation steps are correct, and the final answer follows from the reasoning. Minor wording issues are allowed.
    \item \textbf{Minor Gap}: the overall solution direction is correct, but the trajectory omits some intermediate justifications or contains small gaps that do not invalidate the main reasoning chain.
    \item \textbf{Invalid Reasoning}: the final answer may be correct, but the trajectory contains substantive logical or mathematical errors, such as invalid transformations, unsupported claims, or contradictions.
    \item \textbf{Shortcut}: the trajectory appears to rely on the provided answer rather than deriving it, for example by implicitly working backward from the answer, abruptly introducing the target result without justification, or explicitly revealing the reference answer.
\end{itemize}
We regard \textbf{Valid / Mostly Valid} and \textbf{Minor Gap} as acceptable trajectories, while \textbf{Invalid Reasoning} and \textbf{Shortcut} are treated as problematic trajectories.

\paragraph{Results and analysis.}
Table~\ref{tab:trajectory_quality_audit} shows three notable trends. First, $D_{\text{base}}$ has the highest trajectory quality among the unfiltered data sources, which is expected because these trajectories are obtained through standard rejection sampling without access to answer hints. Second, answer-conditioned trajectories exhibit a higher risk of problematic reasoning before filtering. In particular, $D_{\text{repair}}$ before PURE contains more invalid or shortcut trajectories than $D_{\text{hint}}$ before PURE, reflecting the greater difficulty of the extremely hard problems targeted by local repair.

Third, PURE substantially improves the quality of both hindsight-guided and repaired trajectories. For $D_{\text{hint}}$, the number of problematic trajectories decreases from 5/30 before PURE to 2/30 after PURE. For $D_{\text{repair}}$, the number of problematic trajectories decreases from 8/30 to 3/30. Meanwhile, PURE-rejected trajectories contain a much larger fraction of problematic samples, with 17/30 trajectories judged as either invalid or shortcut-based. This contrast indicates that PURE does not merely remove trajectories at random; instead, it preferentially removes trajectories that are more likely to contain answer-conditioned shortcuts or flawed reasoning.

\paragraph{Implications.}
This audit provides additional evidence that the gains of \textsc{HEAL} are not solely attributable to using the reference answer as a generation hint. While answer conditioning can indeed introduce shortcut rationales, the combination of local repair and PURE filtering yields a substantially cleaner set of synthesized trajectories. Importantly, this audit does not imply that PURE guarantees step-level correctness for every retained trajectory. Rather, it suggests that PURE acts as a useful heuristic for reducing the prevalence of high-risk answer-conditioned rationales, thereby improving the reliability of the resulting distillation data.

\end{document}